\documentclass[conference,compsoc]{IEEEtran}
\usepackage{graphicx}
\usepackage[hidelinks]{hyperref}
\usepackage{cite}

\begin{document}
\title{MLMC: Interactive multi-label multi-classifier evaluation without confusion matrices}

\author{\IEEEauthorblockN{Aleksandar Doknic}
\IEEEauthorblockA{University of Vienna \\ Faculty of Computer Science\\
Email: aleksandar.doknic@univie.ac.at}
\and
\IEEEauthorblockN{Torsten Möller}
\IEEEauthorblockA{University of Vienna \\ Faculty of Computer Science\\
Email: torsten.moeller@univie.ac.at}
\and
\IEEEauthorblockN{Christoph Kralj}
\IEEEauthorblockA{University of Vienna \\ Faculty of Computer Science\\
Email: christoph.kralj@univie.ac.at
}}

\maketitle

\begin{abstract}
Machine learning-based classifiers are commonly evaluated by metrics like accuracy, but deeper analysis is required to understand their strengths and weaknesses. MLMC is a visual exploration tool that tackles the challenge of multi-label classifier comparison and evaluation. It offers a scalable alternative to confusion matrices which are commonly used for such tasks, but don't scale well with a large number of classes or labels. Additionally, MLMC allows users to view classifier performance from an instance perspective, a label perspective, and a classifier perspective. Our user study shows that the techniques implemented by MLMC allow for a powerful multi-label classifier evaluation while preserving user friendliness.
\end{abstract}
\IEEEpeerreviewmaketitle

\begin{figure*}[t!]
 \centering
 \includegraphics[width=\textwidth]{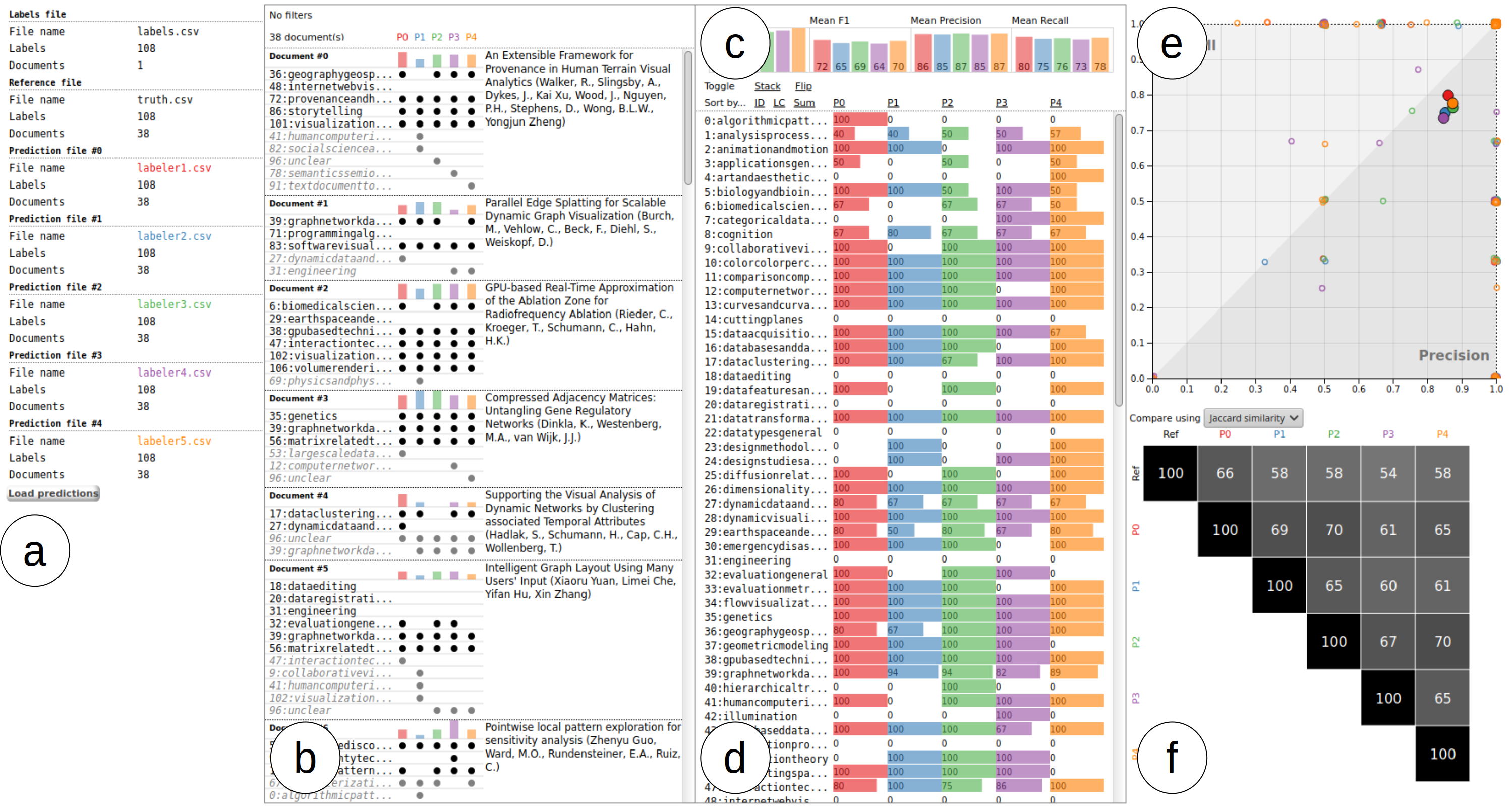}
  \caption{Five classifiers for scientific paper labeling are compared in MLMC: (a) The label names, the ground truth (reference), and the classifier predictions are uploaded by the user. (b) Instance perspective: The paper titles, abstracts, and their label predictions by each classifier are shown in a scrollable list. (c) Classifier perspective: Label cardinality, mean F1 score, mean Precision, and mean Recall are shown for each classifier. (d) Label perspective: F1 scores for each label are represented by bars. (e) Label perspective: Interactive scatterplot shows Precision and Recall for each label and the respective mean centroids. (f) Classifier similarity matrix.}
\label{fig:teaser}
\end{figure*}

\section{Introduction}

Quantitative evaluation measures and confusion matrices are the state-of-the-art method used by the machine learning (ML) community to test and prove the reliability of their classifiers. The problem with global measures
is that they provide a general evaluation and don't tell the classifier engineers \textit{what} the errors are. Visualizations like confusion matrices \cite{usersperspective} are useful for detailed evaluations, but they grow rapidly ($O(n^2)$) with the number of classes. 
A quick view on real-world examples shows that confusion matrices and measures like Precision, Recall, and F1 \cite{evaluatingalgorithms} are commonly used measures in a variety of ML applications. One example is the ML-based classification of protein residues in the context of molecular biology by Mullane et al. \cite{mullane} which was evaluated by using Precision, Recall, and F1 scores. Two other papers, one about music genre classification via convolutional neural networks by Kamtue et al. \cite{kamtue}, and one about blurb labeling via SVM by Bellmann et al. \cite{bellmann} used the same three measures. Another example is ML for melanoma recognition by Codella et al. \cite{codella} which was also evaluated using Precision and Recall, among similar measures. While these are just a few examples, they indicate that a small group of measures is commonly used in a variety of ML applications from different fields. Such measures can be useful, but they merely provide a superficial evaluation (i.e. total performance) when applied beyond binary classification problems and should therefore be augmented by more fine-grained evaluation methods. However, using fine-grained measures to analyze, for example, a classifier with hundreds of classes, would require interpreting at least hundreds of numbers. We suggest that the use of interactive visualization techniques can simplify the process, and enable users to intuitively compare and evaluate classifiers in detail.

In contrast to other approaches, we focus on a more generalized type of classification problem. Classification problems can be separated into three categories: Binary, multi-class, and multi-label. Conventional approaches can be applied for binary cases where only two classes (positive and negative samples) exist. For the multi-class case the question of how to deal with varying error location (e.g. classifier works for one class but not for another) and scalability (i.e. how to treat a large number of classes) arises. Visual techniques addressing the previous issues exist and are discussed in the next section. Multi-label evaluation problems \cite{DBLP:journals/jdwm/TsoumakasK07} are further generalized and more complex because more than one class, now called label, can be assigned to a single instance. 
The issues of error identification, scalability, and finding a method to evaluate a multi-label classifier, are challenges on their own. Additionally, we also wanted to support the comparison of multiple classifiers at the same time to meet the demands of our domain experts. We limit our focus to so-called flat classifiers where each label is on the same semantic level, as opposed to hierarchical classifiers \cite{DBLP:conf/icdm/SunL01} which can be considered a special case.

MLMC was developed in an iterative process inspired by the design study methodology \cite{DBLP:journals/tvcg/SedlmairMM12} together with domain experts, which in our case is defined as someone who uses multi-label classifiers in their research, and visualization experts, which we define as people with experience and prior contributions in the field of visualization and visualization techniques. The domain experts can also be considered the target user group. After initial interviews, prototypes of visualization components were developed, presented to the experts, and discussed. We focused on drafting intuitive interfaces that use familiar aspects but improved and combined them in novel ways that meet our requirements. Colors were consciously used with focus on the classifiers and provide a natural limit to how many classifiers can be evaluated at the same time with our approach. MLMC is therefore particularly suitable for choosing and fine-tuning final multi-label classifier candidates within its scalability boundaries, which are discussed in further sections. The results were evaluated in two small but thorough user studies, one focused on assessing usability, the other on assessing the performance in comparison to a traditional (confusion matrix) approach.

Our contributions can be summarized as:
\begin{itemize}
    \item Analyzing requirements and defining tasks for multilabel classifier evaluation on different levels of granularity.
    \item Implementing an alternative approach to state-of-the-art classifier evaluation like confusion matrices, in particular to evaluate multi-label classifiers.
    \item Based upon our new evaluation approach, we also developed a method of comparing classifier performance on different levels of granularity.
    \item Showing its usefulness by measuring the time, confidence, and accuracy improvements in a user study with two groups.
    \item Evaluating the usability and user friendliness in another, qualitative, user study.
\end{itemize}

\section{Related work} \label{relatedwork}
There are several methods to evaluate classifiers, some of which go deeper into the workings or algorithm of the classifier or are domain-specific. 
Those approaches require tighter integration into the algorithmic aspects. Confusion matrices on the other hand are generated by counting the classification errors (classifier-agnostic).
We also consider the correctness of the final results as if produced by a black box. This section discusses previous visual classifier-agnostic "black box" evaluation tools.

Squares \cite{squares} is a tool for interactive performance analysis of score-based multi-class classifiers. It is designed as an alternative to confusion matrices and relies on juxtaposed color-coded bars, boxes, and histograms for different classes and prediction scores. Accuracy, class-level Precision, and class-level Recall are used for performance estimation. The approach of juxtaposing and color-coding makes it impractical for a larger number of classes and predictors due to increased visual complexity. It is also not suitable for multi-label classification problems. Alsallakh et al. developed the Confusion Wheel \cite{confwheel}, which offers another alternative to the confusion matrix view. It focuses on evaluation of score-based multi-class classifiers. Besides representing the misclassification errors
, similar to a confusion matrix, they also compute Precision-Recall curves, but are limited to showing up to 20 classes at once.

Rivelo \cite{DBLP:conf/sigmod/TamagniniKDB17} offers another interesting approach for visually explaining black box classifiers on an instance-level. The user interface has notable similarities with early versions of the MLMC dashboard. However, their tool is designed for binary classifiers with text data and offers no inter-classifier comparisons. Another instance-level explanation workflow for binary classifiers was developed by Krause et al \cite{DBLP:conf/ieeevast/KrauseDSAB17}. The goals were to assess the accuracy and understand the decisions that the (black box) models for medical applications make. Both approaches rely on using input features of the data for explanations while MLMC focuses on the evaluation aspect and works with any data type.
RuleMatrix \cite{DBLP:journals/tvcg/MingQB19} uses input-output behavior to extract classifier knowledge in shape of a rule-based representation. The authors suggest that it is unclear whether users can get an overall understanding if the list of rules is too long (e.g. over 100). Additionally, it is assumed that classification tasks use fewer than 10 classes.

The What-If-Tool by Wexler et al \cite{DBLP:journals/tvcg/WexlerPBWVW20} allows users to compare performance across models and subsets of input data. Multi-class models are evaluated by Accuracy and confusion matrices while more options are available for binary classification problems. Some of the dataset visualizations have similarities to Google Facets \cite{facets}, another notable data visualization tool. Boxer is a classifier comparison and evaluation tool by Gleicher et al. \cite{boxer}. In contrast to the aforementioned evaluation tools it is also designed to compare multiple classifiers. The distinct feature of Boxer is the use of two selections of subsets of the data. The idea is that subset selections give additional insights, but it requires the user to understand the features and make appropriate data selections.

MyriadCues \cite{DBLP:journals/tvcg/DasguptaWOB20} is another multi-model comparison tool which specifically focuses on model rankings and stability across different quality metrics. MLMC on the other hand is more focused on instance- and label-based comparisons with a limited number of metrics, following the suggestion from Wu and Zhou \cite{DBLP:conf/icml/WuZ17} that some evaluation measures appear effectively redundant (when used for optimizing and evaluating classifiers), and that using instance-wise and label-wise evaluations is more informative.

Finally, we reference UpSet \cite{upset}, a novel technique to visually represent and evaluate aggregations and intersections in data sets. This method inspired some of the visualizations used in MLMC. 

While many of the aforementioned tools include some aspects of MLMC, the distinguishing differences are: our focus on multilabel problems, which are not commonly considered, focus on label scalability (over 100 labels), and support of different datatypes (text, images, audio), evaluating classifications (not classification scores), and considering different levels of granularity (instead of only instances or classes).

\section{Problem analysis}
This section discusses the requirements, tasks, and design goals. In order to establish the requirements we collaborated with two domain experts who were involved in the process of developing multilabel classifiers. The first domain expert was recruited from our own faculty and worked on natural language processing and topic-based labeling of scientific papers while the other domain expert worked for an external institute and developed classifiers for labeling recordings of bird sounds. 

\subsection{Requirements}
We interviewed both of our domain experts to understand the challenges they faced when evaluating their classifiers and identified the following requirements:

\paragraph{R1 Multi-label problems} One challenge was the incompatibility of confusion matrices with multi-label problems. For example, one expert considered creating a confusion matrix by generating all tuples, but questioned the scalability of this approach. Another idea was to only take the most significant label of each instance, and compare it to the most significant label of the ground truth, discarding the other labels in the evaluation process. However, this would not only omit data but also rely on ranked classifier outputs (which only one of our domain experts could produce). The first requirement is therefore support for multi-label problems.

\paragraph{R2 Label-specific evaluations} Both experts relied on global evaluation measures like Accuracy to evaluate the classifiers, but also recognized these measures as insufficient. However, computing and interpreting label-specific measures manually required significantly more effort. Hence, the second requirement is to implement efficient label evaluation methods. 

\paragraph{R3 Comparing predictions on an instance level} Another wish was to go down to the instance level and see what exactly the labeling errors are that each classifier makes and how they differ between classifiers, instead of just looking at the confusion counts. For example, an instance can be assigned two wrong labels, resulting in a certain performance value, but we also want to know \textit{which} labels they are. Understanding the errors can be crucial to finding out how to improve the classifiers.

\paragraph{R4 Data representation} Our domain experts, who work with text data and audio data, requested a convenient way to manually verify the instance level predictions and see if they make sense. They wished for a way to get easy access to representations of the instances, e.g. text files, image files, or audio files, depending on the datatype.

\paragraph{R5 Visual scalability} We consider scalability with respect to the number of classifiers, labels, and instances. Our experts compared 3 or 4 classifiers, with 108 to 1500 labels, and 38 to 3000 instances in the test dataset. It should be noted that, in the case of multi-label classification, even a low number of instances can contain several predictions which can be used to estimate the performance of the classifier. In our case, the data provided by our domain experts showed a label cardinality (average number of labels per instance) ranging from 0.35 to 3.87. For example, if the label cardinality is 3.87, then 100 instances lead to approximately 387 label predictions. Values below 1 occur, for example, if the classification threshold is too high and no predictions are produced for a large number of instances.

\subsection{Tasks} \label{section:tasks}

From the interviews and requirements, we derive two major types of tasks: \textbf{T1} classifier evaluation and \textbf{T2} classifier comparison. Understanding a classifier in detail helps the domain experts make decisions on how to improve the classifier while being able to compare different classifiers enables domain experts to understand the strengths and weaknesses of two or more classifiers. Especially in the case of multi-label this is a non-trivial task due to the fact that different errors can lead to the same performance measures.

\begin{table}[t]
    \caption{The tasks can be structured in two types and three levels of granularity for each type.}
    \centering
    \begin{tabular}{c|c|c|c}
         & Global & Label & Instance \\
        \hline
        Evaluate & T1.1 & T1.2 & T1.3 \\
        Compare & T2.1 & T2.2 & T2.3 / T2.4
    \end{tabular}
    \label{tab:tasks}
\end{table}

\paragraph{T1.1 Evaluating a classifier on a global level}
 
Global classifier performance can be estimated simply by computing the measures which the user deems relevant. 
We do not intend to replace this approach, because we consider it a valid part of the evaluation process, but rather we want to complement it and support the user in uncovering additional problems. For example, 
some users might only evaluate global Accuracy.
This leaves not only a blind spot for class imbalances, but also ignores the perspective on false positive and false negative errors (which are not the same errors but reduce Accuracy in the same way).

\paragraph{T1.2 Evaluating a classifier on a label level}

Evaluating on a label level helps the user decide which parts of the classifier need improvement (e.g. better training data) or if there is a risk of label discrimination (one label suffers from false positives while another suffers from false negatives). However, evaluation can be tricky, because the option of assigning multiple labels to a single instance allows the predictions for an instance to be partly correct and partly incorrect. In fact, having a perfect match between all the ground truth labels and all the prediction labels for a certain instance can become unlikely as the number of possible labels grows. For this reason we focus on the labels and separate them from the instances evaluation. One instance with N possible labels generates N predictions (true positives, false positives, etc.), one for each label, and our evaluation method needs to take this into account.

\paragraph{T1.3 Evaluating a classifier on an instance level}

Evaluating classifiers on a global level and on a label level already covers many aspects that need to be investigated. However, instances should not be ignored for two reasons: First, the performance measures can be deceptive and the same measures produced by different labelings. Because the labels are considered separately, their respective predictions are "thrown in the same bucket" and detached from the instances.  Also, label-based evaluation measures tend to be relative measures and considered with respect to the number of label occurrences in the ground truth. If they do not occur in the ground truth, they cannot be computed. Finally, our domain experts demanded a representation of the data because it allows them to manually evaluate whether the labels in the predictions or in the ground truth make any sense.

\paragraph{T2.1 Comparing classifier performance on a global level}

Comparing global classifier performance helps choosing the most suitable classifier from a number of classifiers. What the best classifier is depends on the evaluation measures and the application domain. While it is important to consider the relative performance for all labels, it is also important to consider their Precision and Recall. Additionally, showing the performance over all instances without considering the labels can provide a complementary view.

\paragraph{T2.2 Comparing classifier performance on a label level}
Two different classifiers could provide vastly difference performance for different labels, but still show very similar global measures. It would be helpful for the user to see (1) if this is even the case and (2) if yes, what exactly are the differences across the labels.

\paragraph{T2.3 Comparing classifier performance on an instance level}
This task compares the performance for each classifier on a single instance, i.e. which labels were assigned to that instance. The idea is to provide an overview over how well a particular instance is classified and what the differences are between classifiers. For example, if 2 out of 3 labels are correct for a particular instance, if 1 out of 3 are correct etc. This task helps the user investigate problematic instances and see if the problem is due to a badly trained classifier, or if it occurs across different classifiers, in which case it could be a data quality problem.

\paragraph{T2.4 Comparing classifier predictions on an instance level}
Following the argument of the previous task, it is also possible that different classifiers label the same instances differently but produce the same or similar performance measures when averaged over all labels. Our domain experts wanted to see how some of the instances are labeled differently by the classifiers in general. Unlike the previous task, the focus is not on the performance, which is always considered with respect to the ground truth, but on the label differences between classifiers.





\subsection{Design goals}
Our goals were to design an interactive visualization tool that (G1) fulfills all requirements above and supports the user in their tasks while being intuitive to use and easy to learn. 
It should also (G2) offer a quick and responsive interface, (G3) allow fast and intuitive processing of tasks, 
and (G4) give the user simple access the actual data (e.g. audio, image, text files) or a suitable representation of the data.
Additionally, (G5) complex menus should be avoided and all data should be shown at the same time where possible, following the vision over cognition approach \cite{DBLP:journals/scholarpedia/Todorovic08}.

\section{MLMC}

MLMC is an interactive web tool that allows the user to evaluate different multi-label classifiers and accomplish the tasks described in section \ref{section:tasks}. Section \ref{section:algorithm} describes the measures that are visualized, and \ref{section:components} describes the visualization components.


\subsection{Algorithmic background} \label{section:algorithm}

\begin{figure}[t]
    \centering
    \includegraphics[width=0.6\linewidth]{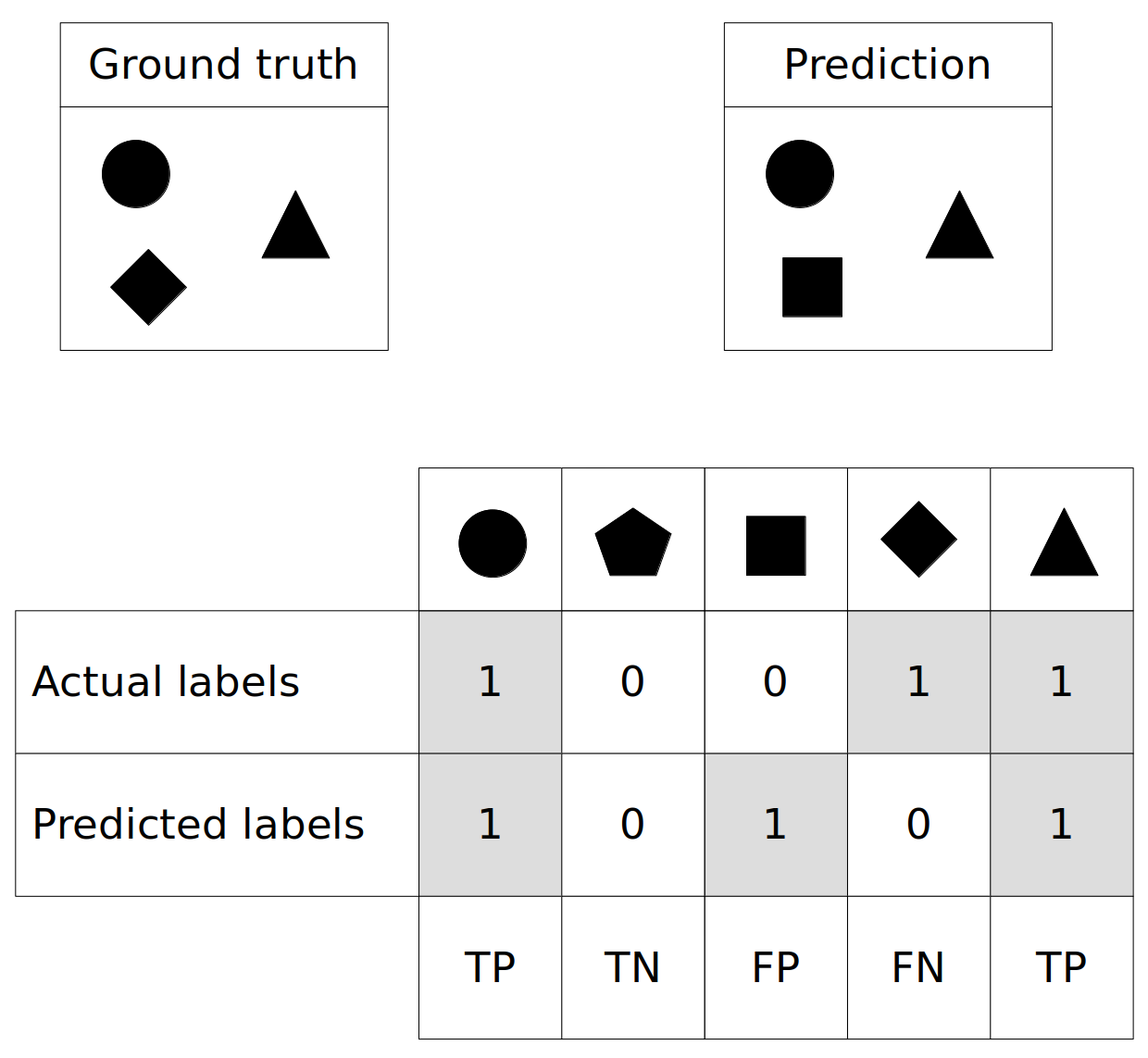}
    \caption{Comparing the ground truth of one instance with a classifier prediction. If the label appears in both, the ground truth and the classifier prediction, it is considered true positive (TP). If it appears in neither, it is considered a true negative (TN). If the label appears in the classifier prediction, but not in the ground truth, it is considered a false positive (FP). Finally, if the label appears in the ground truth but not in the prediction, it is considered a false negative (FN).}
    \label{fig:eval}
\end{figure}

The first step is to count the four types of errors: true positives (TP), false positive (FP), false negatives (FN), and true negatives (TN) as shown in Figure \ref{fig:eval}, for all instances.

Performance measures like Accuracy, Precision, and Recall can be computed by using TP, TN, FP, and FN, which can either be accumulated over all instances, or separately for each label. There is little value in computing these statistical measures for a single instance (sample size of 1) however, which is why we prefer the Jaccard similarity coefficient (eq. \ref{jaccard}) which only considers performance by similarity. We use the Jaccard index $J_i(G_i,P_i)$ to describe how well the set of predicted labels $P_i$ matches the set of ground truth labels $G_i$ for instance $i$:
\begin{equation}
    J_i(G_i,P_i) = \frac{|G_i \cap P_i|}{|G_i \cup P_i|}
    \label{jaccard}
\end{equation}

\textit{Accuracy} is a popular global classifier evaluation measure, because it is intuitive and easy to compute with confusion matrices. However, it does not distinguish between FP and FN errors despite their fundamental differences, e.g. FP can lead to false diagnoses of healthy patients while FN can lead to missing diagnoses on sick patients. Additionally, it adds more weight to classes/labels with many instances more than those with few instances, which can lead to neglected labels.

\textit{Recall} $R_l$ and \textit{Precision} $P_l$ are complementary (label-specific) measures that avoid these problems. $R_l$ close to 1 implies few FN while $P_l$ close to 1 implies few FP. Computing them separately for each label $l$ rules out class imbalances:

\begin{equation}
    R_l = \frac{TP_l}{TP_l+FN_l}
    \label{eq:Recall}
\end{equation}

\begin{equation}
    P_l = \frac{TP_l}{TP_l+FP_l}
    \label{eq:prec}
\end{equation}

The $F_1 score$ is the harmonic mean of both measures and considers both with equal importance.
While label-averaged measures avoid the data imbalance problem, it can still be useful to add an instance-averaged measure for completeness. For this purpose we also use the averaged Jaccard index (eq. \ref{jaccard}).
While there are other suitable measures (e.g. MCC \cite{matthews1975comparison}), we chose Precision, Recall, and F1 score as they are most familiar and useful for our evaluation tasks.

\subsection{Components} \label{section:components}
MLMC consists of four major visualization components, three of which are tightly linked together, allowing to select and highlight a label across the views.

\paragraph{Summary} The Dot Chart shows how each document is labeled by each classifier and allows the user to view the data (i.e. audio, video, or image). The Performance Bars show the F1 score that each classifier achieves (for each label and overall). The Precision-Scatterplot shows the Precision and Recall performance that all labels achieve for each classifier. The similarity matrix shows a quick overview of how similar the classifier outputs are to each other (and to the ground truth) based on Jaccard similarity.

\paragraph{Connection to tasks} Each component is designed for at least one level of granularity from table \ref{tab:tasks}. The Dot Chart is used for instance-level evaluations (T1.3) and comparisons (T2.3 and T2.4), Performance Bars support label-level (T1.2, T2.2) and global-level tasks (T1.1, 2.1), which is also provided by the complementary Precision-Recall Scatterplot. The similarity matrix is purely designed for global-level tasks (T1.1 and T2.1). Colors from ColorBrewer 2.0 \cite{brewer} are used for encoding different classifiers in all visualizations.


\subsubsection{Dot Chart}

\begin{figure}[t]
    \centering
    \includegraphics[width=1.0\linewidth]{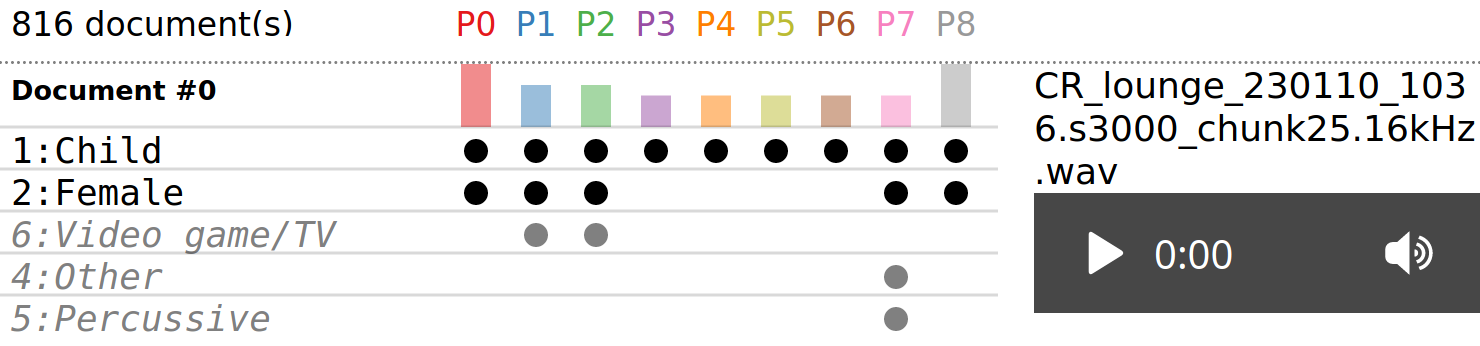}
    \caption{One audio file was labeled by nine different classifiers, which are represented by predictions P0 to P8. The ground truth for this audio file instance is \textit{Child} and \textit{Female}, which describe the recording correctly. Labels which are not in the ground truth are indicated by the grey color. FP errors are therefore made by P1, P2, and P7. P3 to P6 predict \textit{Child} correctly but not \textit{Female} (FN error).}
    \label{fig:dcaselabeldotchart}
\end{figure}

The visualization in figure \ref{fig:dcaselabeldotchart} shows the labelings for a single instance with the respective \textit{document} next to it. A \textit{document} can be an audio file, an image, or a text body, and represents the data that is classified in this particular instance. Task T1.3 is enabled for each classifier separately by the dot visualizations, which expose false positive and false negative errors, as well as the performance bars on top, which describe the matching of ground truth and predictions with the Jaccard similarity coefficient (eq. \ref{jaccard}) from 0\% to 100\%. Values can be seen via mouseover. Task T2.3 and T2.4 are facilitated by an alignment of label dots and bars on the same axis, which allows for a row-wise inter-classifier comparison.
\paragraph{Design decisions} In the early prototypes we realized that alignment plays a significant role as our users needed longer to compare unaligned label predictions. We also considered writing the labels directly instead of using dots, but this requires significantly more space and requires the user to read each label multiple times. Our UpSet-inspired \cite{upset} design allows more compact comparisons visualizations, and considers not only the performance with respect to the ground truth, which is expressed by bars, but also the label predictions themselves (T2.4). This design was justified by the relative simplicity and the suitability for both tasks.

\subsubsection{Performance Bars}

\begin{figure}[t]
    \centering
    \includegraphics[width=1.0\linewidth]{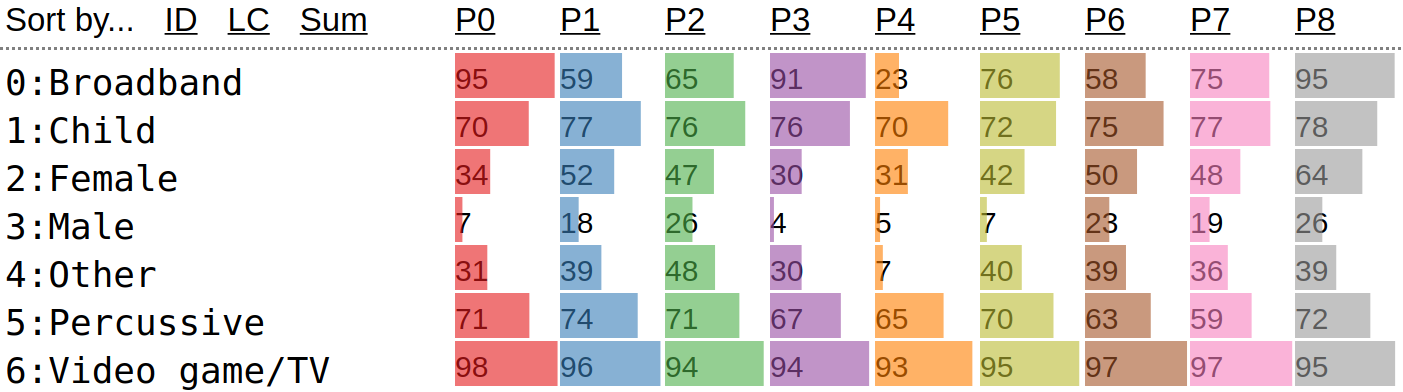}
    \caption{The F1 measure for each label is shown for each classifier within a visually augmented sortable matrix. The columns represent the classifier prediction files while the rows represent the seven different labels that can be assigned to each instance.}
    \label{fig:perfbars1}
\end{figure}

\begin{figure}[t]
    \centering
    \includegraphics[width=1.0\linewidth]{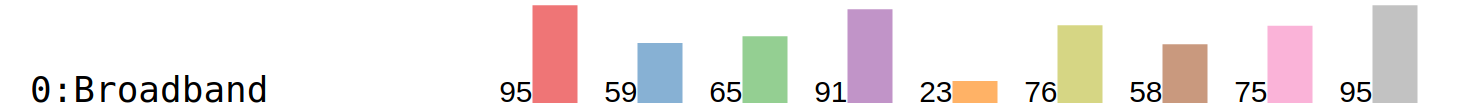}
    \caption{Changing the bar orientation simplifies comparison across different classifiers for T2.2.}
    \label{fig:perfbars2}
\end{figure}

\begin{figure}[t]
    \centering
    \includegraphics[width=1.0\linewidth]{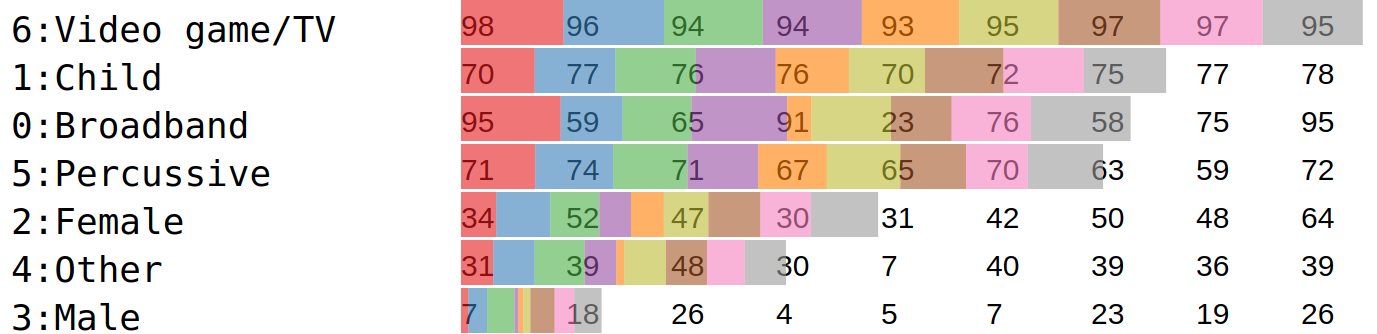}
    \caption{Sorting and stacking the bars reveals label performance accumulated over all classifiers in descending order.}
    \label{fig:perfbars3}
\end{figure}

Unlike the Dot Chart, which considers only one instances at a time, the performance bars show the performance over all instances, but consider each label separately, allowing for label-wise evaluation (T1.2) and comparison (T.2.2). The bars in figure \ref{fig:perfbars1} represent the label-specific F-score $F_1 = 2(P_l R_l)/(P_l+R_l)$ with $R_l$ and $P_l$ from eq. \ref{eq:Recall} and eq. \ref{eq:prec}. Labels that generate neither true positives nor any errors are considered to have an F-score of 100\% due to the fact that they at least represent true negatives.

\paragraph{Design decisions}
The visualization is based on superposing a score matrix with interactive bars to combine both aspects, numerical accuracy and visual interpretability. Alternative encodings such as heatmaps were considered, but we found that bars are more flexible towards user interaction, and lengths are more accurately perceived by humans than differences in brightness according to Steven's Psychophysical Power Law \cite{stevens1986psychophysics}. In their default orientation (figure \ref{fig:perfbars1}) the bars are aligned to facilitate comparison of label performance per classifier for task T1.2. Sorting by ID, label cardinality (avg number of labels per instance), F1 score per classifier and across classifiers, simplifies visual analysis, especially if the number of labels is larger than in this example. Additionally, users can flip the bars to simplify comparison of label performance across multiple classifiers for task T2.2 as shown in figure \ref{fig:perfbars2}. Stacking the bars like in figure \ref{fig:perfbars3} is another visual analysis tool provided to the user, which helps identifying particularly difficult labels across multiple classifiers. The design was justified by the familiarity and interpretability, as well as flexibility (e.g. simple sorting and flipping) which allows different points of view.

\begin{figure}[t]
    \centering
    \includegraphics[width=1.0\linewidth]{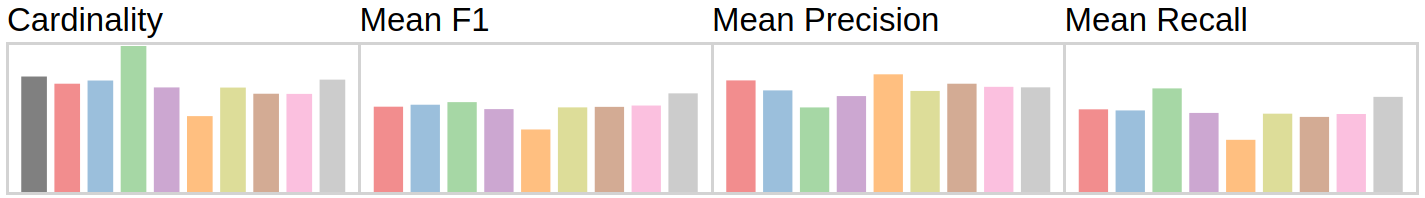}
    \caption{Vertical bars on top show measures averaged across all labels for each classifier. Cardinality refers to the average number of predicted labels per instance for the respective classifier (ground truth in dark grey).}
    \label{fig:perfbars4}
\end{figure}

Finally, summary measures averaged across all labels can be seen in figure \ref{fig:perfbars4}, providing one of the ways for the user to globally evaluate (T1.1) and compare (T2.1) classifier performance. The file names and numerical values can be revealed via mouseover.

\subsubsection{Precision-Recall Scatterplot}

\begin{figure}[t]
    \centering
    \includegraphics[width=0.5\linewidth]{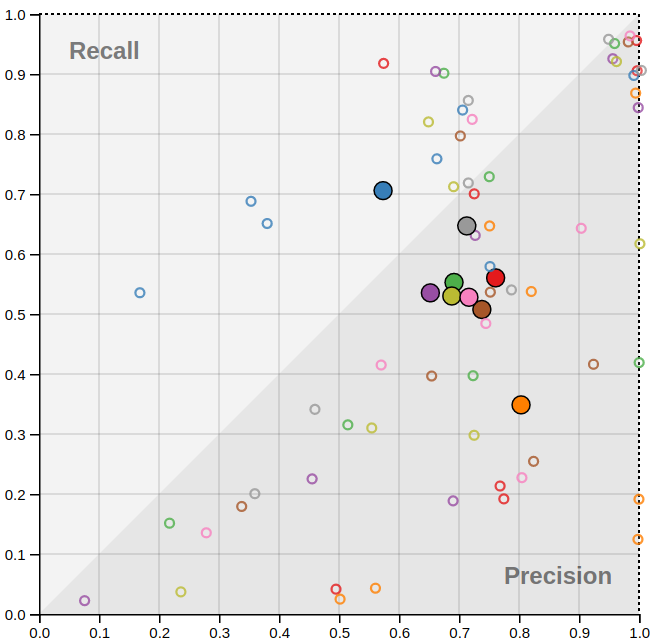}
    \caption{The Precision-Recall scatterplot shows the error bias for each classifier (opaque points) and each label (small circles). Most labels are in the lower triangle, indicating better Precision than Recall, i.e. not recognizing a label (false negative) is more common than assigning a wrong label (false positive).}
    \label{fig:splot1}
\end{figure}

\begin{figure}[t]
    \centering
    \includegraphics[width=1.0\linewidth]{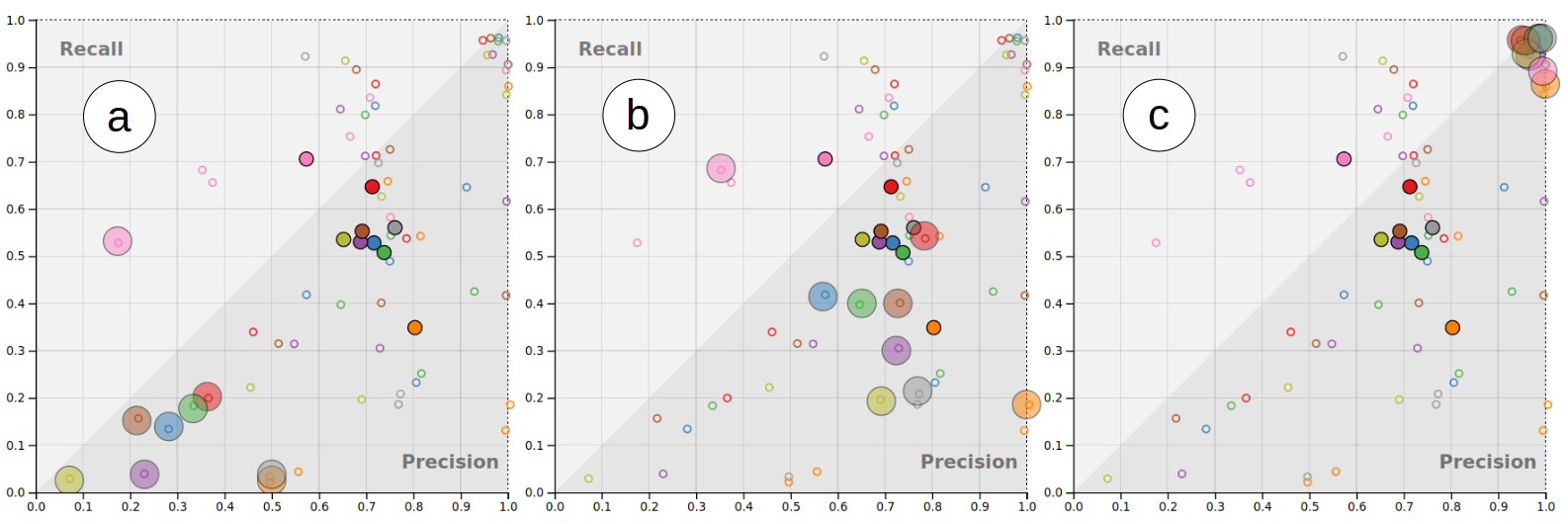}
    \caption{The same label is highlighted (large circles) for different labels: (a) Male, (b) Female, (c) Videogame/TV.}
    \label{fig:splot2}
\end{figure}

The performance measures only consider F1, but simply changing the bars to either represent Precision or Recall would hide the balance between Precision and Recall. Precision and Recall are shown in form of a scatterplot next to the performance bars visualization. The scatterplot is heavily linked to the performance bars. Every F1 bar in figure \ref{fig:perfbars1} is linked to the respective point in the scatterplot in figure \ref{fig:perfbars4} which allows a quick view on Precision and Recall. It is also possible to highlight a label for all classifiers as shown in figure \ref{fig:splot2}.

\begin{figure}[t]
    \centering
    \includegraphics[width=0.5\linewidth]{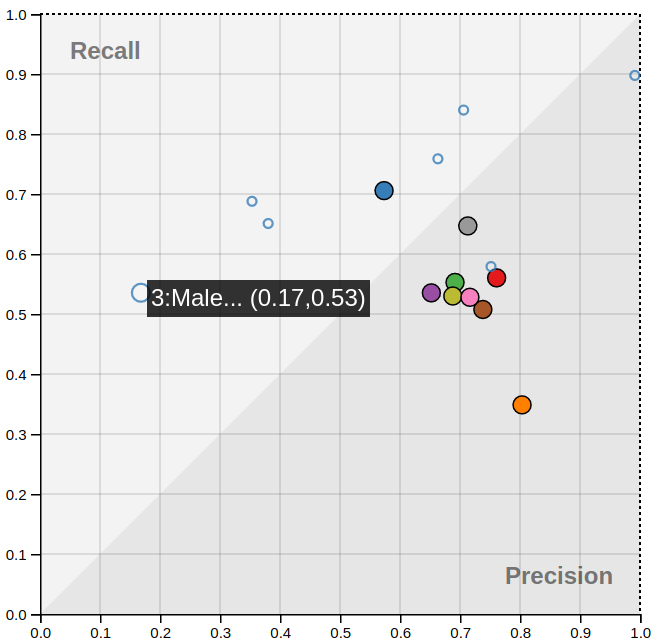}
    \caption{Hovering over a label shows further details: the label ID and the first letters as well as the  Precision and Recall values. To reduce cluttering, hovering over a blue label hides all non-blue labels (from other classifiers).}
    \label{fig:splot3}
\end{figure}

Points at the diagonal border indicate balance between both error types. For a perfect classifier with an F1 score of 100\% all points would converge to the top right corner. Overlaps are avoided by adding a small random disposition to each point. Additionally, hovering over a label point shows the label name, numerical Precision and Recall, and hides all points from other classifiers, see \ref{fig:splot3}. Clicking on a point highlights the specific label in the Performance Bars view.

Similarly to the Performance Bars, the user can explore label-level performance (T1.2), e.g. figure \ref{fig:splot3}, for multiple classifiers (T2.2), e.g. figure \ref{fig:splot2}, as well as global measures, which are represented by the opaque Precision-Recall centroids (figure \ref{fig:splot1}, \ref{fig:splot2}, \ref{fig:splot3}).

\paragraph{Design decisions}
While there are drawbacks to using scatterplots, specifically the fact that it is not obvious which points belong to which label, this drawback is compensated by the Performance Bars and the interaction between both visualizations. Precision-Recall bias plays an important role in classifier evaluation, both, globally and for each label, and the scatterplot with the diagonal border specifically designed to address this part.

\subsubsection{Similarity matrix}

\begin{figure}[t]
    \centering
    \includegraphics[width=0.5\linewidth]{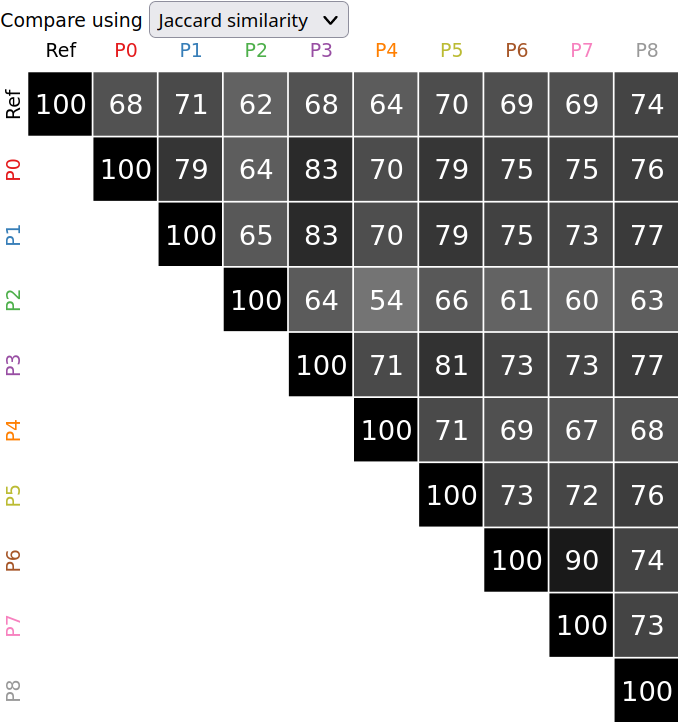}
    \caption{Similarity matrix shows the instance-averaged Jaccard similarity between classifiers (P0-P8) and ground truth (Ref).}
    \label{fig:sim}
\end{figure}

The Similarity Matrix offers a complementary view to the previously mentioned label-based evaluation methods. Classifier performance (T1.1 and T2.1) can be evaluated and compared by interpreting the first row (figure \ref{fig:sim}), since the first row represents the similarity to the reference data (ground truth) itself while the other rows compute prediction similarity between classifiers. The numbers represent the averaged Jaccard index of the bars in the Dot Chart (see figure \ref{fig:dcaselabeldotchart}) over all instances. These values can be considered complementary because they are \textit{not} relative to the labels, i.e. they do not take label imbalance into account, in other words, labels which occur more frequently in the ground truth have a larger impact. Figure \ref{fig:sim} shows that the predictions made by classifier P6 and P7 have 90\% overlap. In this particular case, both classifiers were created by the same person.

\paragraph{Design decisions} The choice of using a similarity matrix can be attributed to user familiarity with this type of visualizations for distances (e.g. distance matrices), as well as the fact that it can be quickly interpreted and structured in a logical way, unlike, for example, 2-dimensional embeddings with their own distance metrics. Using luminosity to encode the similarity value is justified by the fact that the focus lie on relative similarities (e.g. which classifiers are most similar to each other), and not exact values, unlike the Performance Bars, where both aspects can be important.

\subsubsection{Interactions}
Selecting a label in one visualization highlights it in all visualizations. Additionally, all instances that do not include the selected label are removed from the Dot Chart to allow the user to focus on the relevant instances.

\section{Use cases}
MLMC has been tested with five datasets, two of which are publicly available. This section describes three use cases with three different datatypes. The difference between data types is the data representation in the Dot Charts.

\subsection{Audio: DCASE 2016} \label{section:dcase}

The IEEE AASP Challenge on Detection and Classification of Acoustic Scenes and Events \cite{Mesaros2018_TASLP}, in short DCASE, is a yearly competition organized under supervision of the Audio and Acoustic Signal Processing technical committee of the IEEE Signal Processing Society. The (publicly available) submissions for the fourth task contained score-based label predictions, with the scores ranging from 0 to 1. For the evaluation with MLMC a threshold of 0.5 has been chosen. 
We evaluate the resulting classifications for 816 audio files with 4 seconds length each. The labels are: Broadband, Child, Female, Male, Other, Percussive, and Video game/TV.

All figures from section \ref{section:components} show this dataset. From a global perspective, all but one classifier in figure \ref{fig:splot1} are biased towards Precision. Classifier P8 (grey) has the highest mean F1 score as shown in figure \ref{fig:perfbars4}, as well as the highest Jaccard similarity towards the ground truth (74\%) as shown in figure \ref{fig:sim}. From a label perspective, it can be seen that the label \textit{Video game/TV} has the highest F1 score over all classifiers, see figure \ref{fig:perfbars1} and \ref{fig:perfbars3}. The labelings for one of the instances are shown in figure \ref{fig:dcaselabeldotchart}: We see that P8, the classifier with the highest global performance, correctly predicted the labels for this instance.

Evaluating audio classifiers suffers from one drawback: Unlike image files or text files, the user has to play and listen to each audio file separately in order to evaluate it manually. Methods to visualize audio files in this context could be considered in future works.

\subsection{Image: Flickr8k} \label{section:flickr}
The Flickr8k dataset \cite{flickr} is a publicly available diverse set of 8000 pictures. It mostly depicts humans or dogs participating in an activity. 256 annotated pictures with 224 labels were taken from this dataset and a reference based on the included annotations was created. The agreement of crowd voters (this information was provided in the original dataset, each voter said either yes or no to a label) for each label ranged from 0 (no one agrees that the annotation fits the picture) to 1 (everyone agrees). Prediction files were then generated for three different thresholds: 0.25, 0.5, and 0.9. If the agreement is larger than the threshold, the label is assigned to the picture.

\begin{figure}[t]
    \centering
    \includegraphics[width=1.0\linewidth]{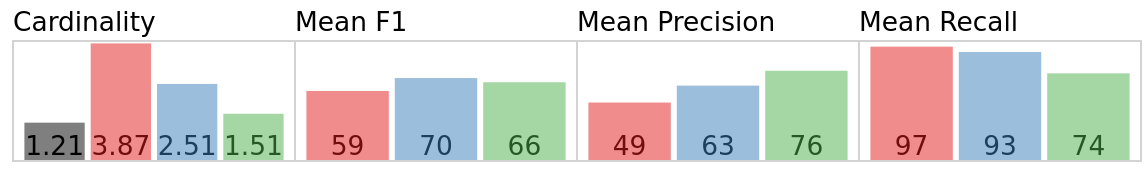}
    \caption{Flickr8k dataset \cite{flickr}. Red bars represent a classifier with a threshold of 0.25, blue represents 0.5, and green represents 0.9.}
    \label{fig:flickr1}
\end{figure}

\begin{figure}[t]
    \centering
    \includegraphics[width=0.5\linewidth]{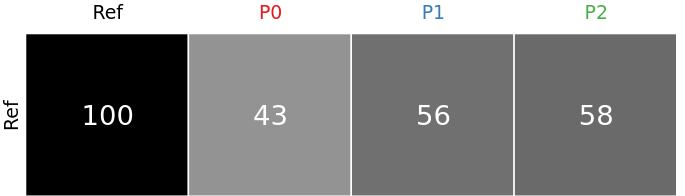}
    \caption{First row of the similarity matrix of the Flickr8k dataset \cite{flickr}.}
    \label{fig:flickr5}
\end{figure}

\begin{figure}[t]
    \centering
    \includegraphics[width=0.5\linewidth]{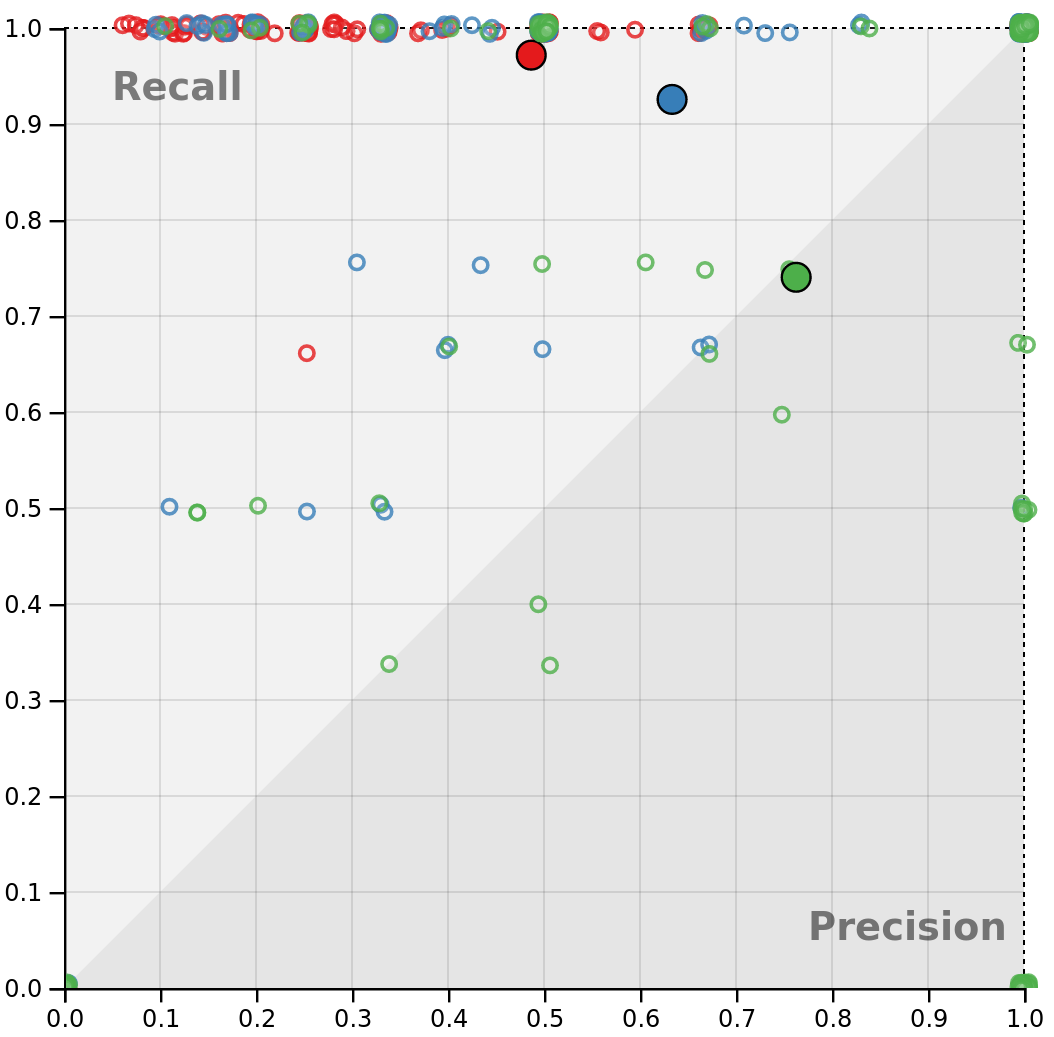}
    \caption{Flickr8k dataset \cite{flickr} scatterplot shows that two classifiers have a preference for Recall.}
    \label{fig:flickr2}
\end{figure}

Figure \ref{fig:flickr1} shows that the Precision and Recall behave as expected: A lower threshold leads to a higher Recall but lower Precision while a higher threshold leads to a lower Recall but higher Precision. The large cardinality values imply that the instances receive too many labels compared to the ground truth, which on average has 1.21 labels per instance. A threshold of 0.5 leads to the highest F-score (70\%), but assigns more than twice as many labels to each instance than the ground truth. The similarity matrix (figure \ref{fig:flickr5}) on the other hand shows a slightly higher similarity (58\%) towards the ground truth for the 0.9 threshold than for the 0.5 threshold (56\%). In figure \ref{fig:flickr2} we can see that the threshold of 0.9 leads to the most error-balanced classification behavior among the three. The question of what the best classifier is clearly depends on the application and what errors are more acceptable.

\begin{figure}[t]
    \centering
    \includegraphics[width=1.0\linewidth]{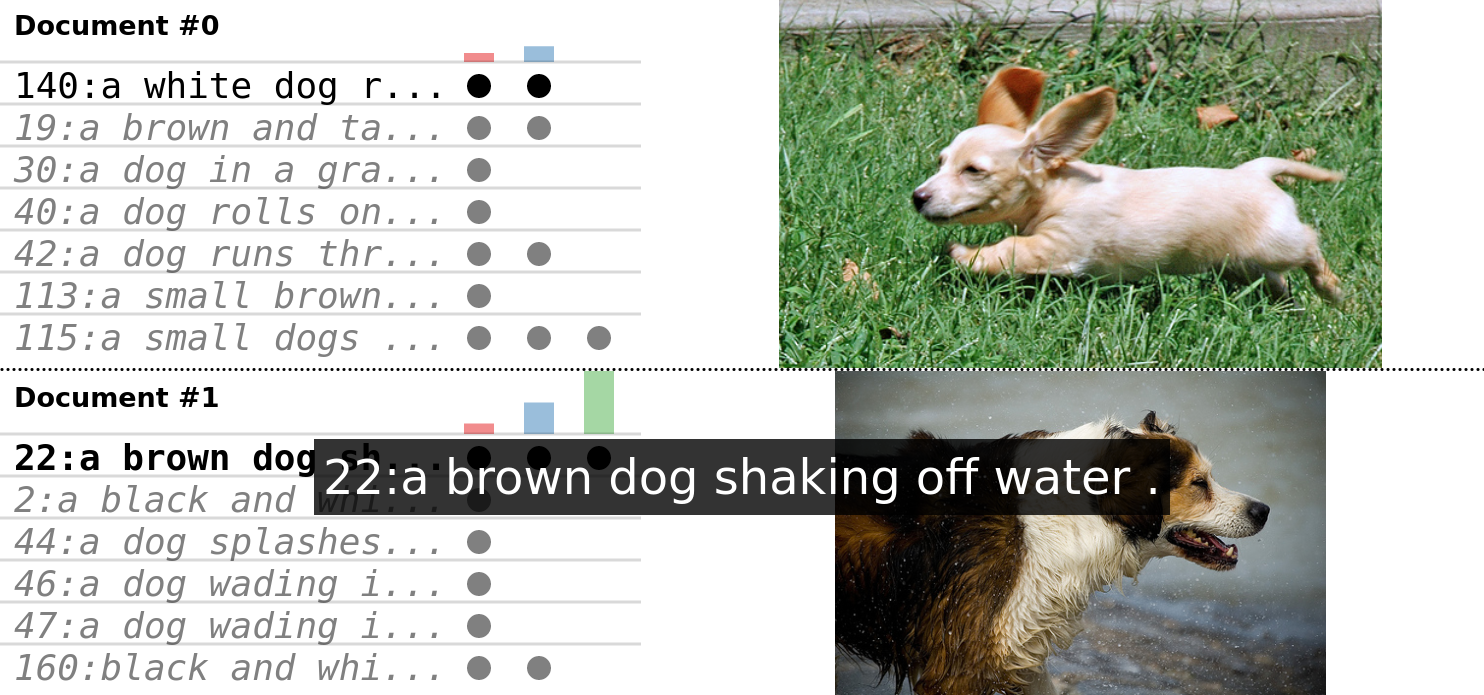}
    \caption{Two Dot Charts from the Flickr8k dataset \cite{flickr}.}
    \label{fig:flickr4}
\end{figure}

The Dot Charts in figure \ref{fig:flickr4} show the over-labeling problems in greater detail, and reveal how a high threshold leads to more false negatives for particular images. 

\subsection{Text: Paper labeling} \label{section:paperlabeling}
\begin{figure}[t]
    \centering
    \includegraphics[width=1.0\linewidth]{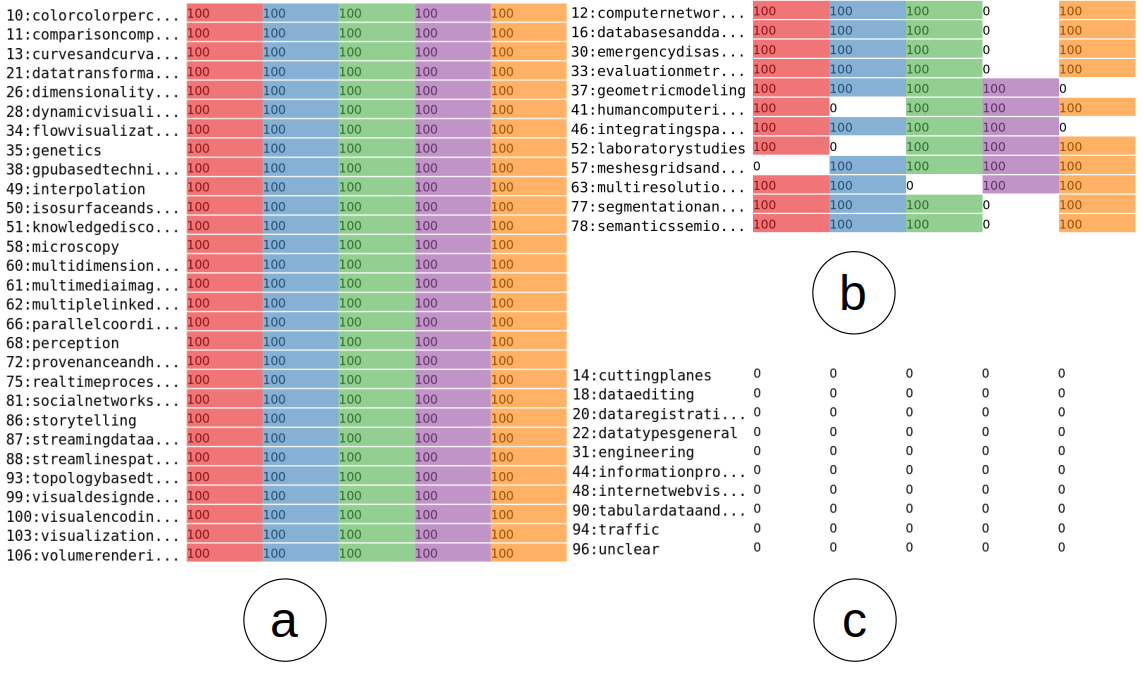}
    \caption{Sorting by the total sum of F1 scores reveals how well each classifier recognizes each of the labels: (a) work for all classifiers, (b) work for some but not for others, and (c) have no true positives for any classifier.}
    \label{fig:paperlabeling}
\end{figure}

Figure \ref{fig:teaser} shows a text body classification dataset from one of our domain experts. The dataset shows similar measures and classification behavior across all five classifiers. From the global measures in figure \ref{fig:teaser} (c), the scatterplot in (e), and the similarity meaures in (f), we can assume that P0 provides the best classification performance overall. At this point it makes sense to investigate which labels are commonly misclassified according to (d) and the related documents (d). One method to do this is to sort by total label performance and take a look at the 0 scores as shown in figure \ref{fig:paperlabeling}. We can see that ten of the labels are never recognized correctly while producing false negative and/or false positive errors.

\section{Usability study}
For the usability study we recruited six users with computer science background, five male and one undeclared, all in the age group of 21 to 30 years. In a self-assessment question the users described their experience in the field of neural networks and classification as ranging from 1 (\textit{Never heard of it}) to 5 (\textit{I know a lot about it}). All options were selected by the users, with 4 selected twice.

The user received a short explanation for each visualization part of the software before they had to solve tasks for two use cases while thinking aloud. This approach was inspired by the Thinking-aloud method \cite{thinkingaloud}. After the interactive part of the study, the user received a web link to a three part online questionnaire consisting of general questions, selected ISONORM 9241/110 questions, and System Usability Scale (SUS) \cite{sus} questions.

This study design was influenced by the COVID-19 crisis and participants connected via a remote desktop tool (TeamViewer) to work with MLMC. The Flickr8k dataset from section \ref{section:flickr} and the paper labeling dataset from section \ref{section:paperlabeling} were pre-loaded and prepared for usage on a test PC. After a short introduction, the users were verbally asked the same questions for both datasets and were given one hour in total to learn how to use the tool and answer four questions. They were asked to describe the differences between the predictors, their personal preference, an example for a good label and an example for a problematic label. The questions were not supposed to propose a challenge and there were no true or false answers. The goal was to get an informal assessment, to see if the task and software functions are understood, and to allow the test users to gain experience before they fill out the questionnaire later on.

\subsection{Observations} One observation during the task solving was that some users like to scroll around, but sometimes forget about the sorting and stacking buttons unless they're told how to use them. The label \textit{unclear} in the paper labeling dataset was identified as problematic by at least two users, which is correct, because it's a data artifact. One user noted that some labels might be problematic because they were either too general, or too specific for the document. It was also suggested that label overlaps could be considered, for example the semantic similarity of the labels to each other. Another suggestions was to sort or filter by predictor disagreement in order to emphasize only the documents and labels where actual differences can be seen. One user complained that the scroll bar would jump back whenever the mouse is moved away from the abstract text area in the text use case. Overall the tasks were solved to a satisfactory degree and the users seemed to understand how to use the tool on a basic level. 
After the testing the tool, every user filled out an online questionnaire.

\subsection{Questionnaire results} The mean value of the points assigned to ISONORM questions was 5.16 out of 6 (86\%) averaged for all questions for all test users.  The answers to the SUS questionnaire, computed by the original method, show a score of 83 out of 100 points with the lowest being 77.5 and the highest being 90. While the results are generally satisfying, there were two questions that were rated worse than the others: The statement \textit{I felt very confident using the system} received an average score of 2.5/4 (higher is better), and \textit{I needed to learn a lot of things before I could get going with this system} received a score of 1.17/4 (lower is better). This indicates that there was still some degree of insecurity when using the software. No significant usability score difference between users with lower reported experience and users with higher reported experience was observed.

\section{Performance study}

\begin{table*}[t]
    \caption{Performance study results. Group G decides the order of evaluation methods. Classifier knowledge (Class.) and visualization knowledge (Vis.) is self-assessed on a scale from 1 to 5. Participants were also asked how often they use spreadsheet applications (Sp.s.). The results include the total time for answering all questions T, the average time per task T/task, the number of correct responses Res., and the average confidence C. that participants had in their answers.}
    \centering
    
    \begin{tabular}{c|c|c|c|c|c|c||c|c|c|c||c|c|c|c}
        \hline
        \multicolumn{7}{c||}{Participant data} & \multicolumn{4}{c||}{Confusion Matrix} &  \multicolumn{4}{c}{MLMC} \\
        G & Age & BG & Edu. & Class. & Vis. & Sp.s. & T & T/task & Res. & C. & T & T/task & Res. & C.\\
        \hline
        \hline
        1 & 21-30  & Chemistry & Master & 1.5 & 4 & monthly & 25:32 & 05:06 & 2/5 & 3.0 & 01:45 & 00:21 & 5/5 & 4.1\\
        1 & 21-30 & CS & Bachelor & 3 & 5 & weekly & 15:04 & 03:01 & 2/5 & 3.8 & 02:41 & 00:32 & 5/5 & 4.6\\
        1 & 21-30 & CS & Master & 4 & 4 & yearly & 14:38 & 02:56 & 5/5 & 3.4 &04:26 & 00:53 & 4/5 & 4.4 \\
        2 & 31-40 & CS & Master & 2 & 4.5 & monthly & 37:37 & 07:31 & 3/5 & 1.9 & 02:33 & 00:31 & 5/5 & 4.6\\
        2 & 21-30 & Math & Master & 3 & 3 & weekly & 12:04 & 02:25 & 4/5 & 2.6 & 01:48 & 00:22 & 5/5 & 4.4\\
        2 & 21-30 & Physics & Master & 4 & 4 & daily & 33:07 & 06:37 & 3/5 & 2.0 & 03:43 & 00:45 & 5/5 & 4.6\\
        \hline
    \end{tabular}

    \label{tab:perfresults}
\end{table*}

Two groups of test users were asked to analyze classifier output and answer five questions.  
Even though there are interactive visualization tools that are designed for similar purposes (as discussed in section \ref{relatedwork}), none of them seems widespread enough to focus on a for a comparison.
Summative quality measures and confusion matrices are still preferred evaluation methods. We therefore decided to compare MLMC to confusion matrices spreadsheets, which also gives users the flexibility to easily compute performance measures if necessary. We tried to recreate a realistic situation, i.e. participants were allowed to access external resources and edit the provided files.

\paragraph{Participants}
Test users were recruited from the faculties of computer science, chemistry, and mathematics, and consisted of 6 male master-level students and doctoral candidates between 24 and 38 years. The users were split into two groups. Both groups received the same two datasets and viewed them with the same two methods, but in different order.

\paragraph{Setup}
 Group 1 first received a Google spreadsheet containing a confusion matrix with cells that were highlighted according to the values (white for zeros, light grey to dark grey for everything else). Column sums and row sums were pre-computed, and diagonal elements were extracted to an extra column. Group 2 first received the MLMC web tool with the data pre-loaded. 
We decided that, even though instance-level evaluation is an advantage of MLMC, it should not be included in this study for comparison reasons. Removing the document dot chart should not affect usability too much because usability was already shown to be high in the last section which included the dot chart.
We also decided to use the DCASE dataset from section \ref{section:dcase} which has a reasonable number of labels (7) and instances (816), and limited the evaluation to 5 classifiers instead of 9 like in the original dataset. 

Multi-class confusion matrices represent only one class per row and per column. We therefore defined all possible tuples of labels as \textit{classes}. However, we cannot always represent all possible classes (for example, for $N$ labels there would be $2^N$ combinations in general) so we prefer to only show those that actually occur in the ground truth or in the predictions. The size of the matrix depends on the number of unique sets of labels in the ground truth and prediction data. Hence, the confusion matrix has to be pre-computed in consideration of all label tuples that occur in all predictions, otherwise the dimensions of the matrix would vary for each classifier. This reduces the number of classes in our case to 80 and the matrix dimensions to 80x80.

We asked the user to identify the best performing classifier out of five (Q1), to find the most similar classifier to a given classifier (Q2), to describe if a specific classifier has more false positive or more false negative errors (Q3), and to find the label with the least (Q4) and most (Q5) relative prediction errors for a specific classifier. 

\paragraph{Results}

The summarized results are shown in Table \ref{tab:perfresults}.
We notice that the first question was mostly answered correctly with the help of the confusion matrix, with 5/6 users giving the expected answer, while the remaining questions received fewer correct responses (4/6, 4/6, 4/6 and 2/6). In Q3 participants noticed the triangular shape of the matrix and understood the meaning but struggled with the interpretation. Low scores in Q4 and Q5 could be explained by the fact that the labels are spread over several tuples and require additional grouping to evaluate, which requires some experience. One participant with (self-assessed) classifier knowledge of 4/5 answered every question correctly only using the confusion matrix. On the other hand, answers based on the MLMC tool were generally correct, with the exception of Q2 (5/6) which had one unexpected answer. In this particular case the similarity matrix was overlooked and the answer was given based on the similarity in label-specific F1 performance, which could be considered a valid approach. While it can be argued that other answers could also be valid by measures or standards that we do not consider here, we observe that the answers provided when using MLMC are more consistent, and time and confidences also improve significantly, in comparison to a confusion matrix spreadsheet.

\section{Discussion}
The qualitative user study indicates that the users were quickly able to learn how to use the tool regardless of their backgrounds (design goal G1), even though there is room for improvement with an SUS score of 83. The user interface implemented in JavaScript with D3 \cite{d3} has no noticeable loading times for the tested data. It runs smoothly in every tested situation and browser, meeting goal G2. Similarly, goal G3 was reached according to the performance user study which showed that certain tasks can be solved relatively quickly compared to not having access to the tool, see table \ref{tab:perfresults}. Goal G4 was achieved by integrating the instance representation (image, audio, or text file) into the dashboard (e.g. figure \ref{fig:teaser}, \ref{fig:dcaselabeldotchart}, \ref{fig:flickr4}), where in the case of audio the files are only loaded and played on demand for performance reasons. The dashboard design also avoids nested menus while still being capable of fulfilling all tasks, therefore achieving goal G5.

Our qualitative user study motivated users to come up with new ideas and potential improvements which could be considered for future versions with different tasks. We have shown that MLMC can be used for different types of data, but for a more enhanced exploration and interpretability it would be interesting to explore subsets of the data like Boxer \cite{boxer}. The Flickr8k dataset \ref{section:flickr} compares crowd-sourced predictions based on different scoring thresholds. However, the threshold can be set at any value between 0 and 1 and it would be interesting to see the classification behavior across the threshold, which is for example considered by ROC \cite{fawcett2006introduction} plots.

Scalability is another important aspect that deserves to be discussed. The number of classes which can be compared simultaneously is limited by the space and the number of colors which can be clearly distinguished. For our use cases it works well for up to 9 classes, see figure \ref{fig:perfbars1}. While MLMC was able able to work with up to 1500 labels, we only tested it up to 224 labels in our user study. At least a few hundred labels can be analyzed if sorting mechanisms are applied (e.g. see figure \ref{fig:paperlabeling}). One of the biggest limitations of MLMC is the instance scalability. Depending on the label cardinality, a few thousand instances can be loaded and filtered by their labels, and the user can evaluate specific instances individually. For tens of thousands or hundreds of thousand instances, it would appear more reasonable to focus on the label-based or global measures, or reduce the test data to a smaller subset of instances, which involves preprocessing.

Future projects could focus on the development of a tool which allows the quick comparison of hundreds or thousands of auto-generated classifier models in order to find the best ones, which can then be compared in detail. The current MLMC prototype could then be considered the detailed view. It would also be interesting to consider score-based classification and ranked classification in future projects.

\section{Conclusion}
Instead of focusing on single visualization techniques, like bar charts or parallel coordinates, we offer a solution for a practical problem that is becoming more common as ML becomes a part of the standard toolkit in computer science.

We present one of the first visualization papers that discusses multi-label classification problems, in contrast to previous works, which mostly focus on multi-class but single-label problems. We present some of the most common tasks and challenges in the context of multi-label evaluation and present possible solutions with the example of three use cases in three very different domains. Our user studies show excellent results, validate our approach, and confirm that our tool is easy and intuitive to use with potential for a variety of application areas. We also showed that less time that is required to solve common classifier evaluation tasks, received more consistent responses, and increased user confidence significantly.
\bibliographystyle{IEEEtran}
\bibliography{MCV_manuscript}

\end{document}